\documentclass[Afour,sagev,times]{sagej}

\usepackage[T1]{fontenc}
\usepackage[utf8]{inputenc}
\usepackage{listings} 
\usepackage{moreverb,url}
\usepackage{float}
\usepackage[colorlinks,bookmarksopen,bookmarksnumbered,citecolor=red,urlcolor=red]{hyperref}


\newcommand\BibTeX{{\rmfamily B\kern-.05em \textsc{i\kern-.025em b}\kern-.08em
T\kern-.1667em\lower.7ex\hbox{E}\kern-.125emX}}

\def\volumeyear{2026}

\begin{document}

\setcitestyle{numbers,square}


\title{Characterising AI Models for Cataloguing}

\author{Miguel Arana-Catania\affilnum{1} and Neil Jefferies\affilnum{1}}

\affiliation{\affilnum{1}University of Oxford, UK}

\corrauth{Miguel Arana-Catania, University of Oxford, Broad Street, Oxford, OX1 3BG, UK}

\email{humd0244@ox.ac.uk}

\begin{abstract}
The creation of digital collections involves not only the digitisation of content, but also the creation of catalogue records for it. This often-overlooked task requires slow and costly expert manual work. In this project, we have evaluated the application of AI models to this task, comparing different implementations and models. This work includes a qualitative and quantitative evaluation of the experiments carried out, as well as recommendations on the use of AI models that go beyond the specific use case.
\end{abstract}

\keywords{cataloguing, llms, multimodal, digital collections}

\maketitle

\section{Introduction}
The broad objective of this work at the Bodleian Libraries was to build in-house understanding and expertise in AI tools based on Large Language Models (LLMs), with a particular focus on cataloguing-related tasks. In this project, this has been approached from two main angles:
\begin{enumerate}
    \item Understand the performance of various AI and other automated tools in creating or extracting metadata from a variety of sources, in comparison with the benchmark records from the previous activity.
    \item Understand the broader behaviour of AI tools that may influence future deployment, such as their reliability, consistency, and dependency on prompt construction.
\end{enumerate}

AI technologies have been applied in diverse ways to the task of library cataloguing and classification. \cite{mahmud2024ai, engel2025artificial} provide comprehensive reviews of these explorations. In this regard, prior to our project, \cite{osmani2025machine} evaluated the generation of subject headings using human-generated metadata from the works as input; \cite{martorana2024zero} used LLMs for topic classification using a controlled vocabulary; \cite{gonzalez2025metadata} worked with university theses, using AI to identify the ‘college’ field in the metadata record; \cite{golub2024automated} applied traditional machine learning techniques to the generation of Dewey Decimal Classification classes; \cite{lorang2020digital} used machine learning to classify documents based on their images; \cite{busch2025exploring} fine-tuned LLMs to generate specific metadata fields such as title, language or date; and \cite{song2026comparative} used LLMs to generate Library of Congress Classification classes.

Whilst these previous works offer interesting insights into the use of AI for specific tasks within the cataloguing process, our project aims for a much more ambitious goal. In this work, we employ an end-to-end approach using AI models for the entire cataloguing process. Our experiments employ scanned images of the content as input and produce the complete metadata record as output, in the appropriate encoding format (MARC, BIBTEX, JSON), which includes not only the content of the relevant fields but also the correct formatting code.

\section{Methodology}

In this section, we describe the methodology used in this article. In brief, this has involved applying various LLMs to a collection of digitised document images with the aim of automating the creation of catalogue records for these documents.

Below, we present details of the data used, as well as the experimental and evaluation methodology.

\subsection{Dataset}
The dataset used in this project was drawn from the Bodleian Libraries' Global Dissertations collection that had been newly scanned. This, mainly 16th \& 17th Century European, dataset had the advantage that it was very unlikely that the texts had previously been made available online, so the AI algorithms under test would not have been trained on them. The results obtained would thus be an accurate reflection of cataloguing performance rather than the retrieval of information already embedded in the AI model. A card catalogue for the collection was also scanned and used to provide a human-generated ground truth for evaluation purposes.  

The corpus consisted of 87 dissertations, with a median of 16 page images per item. The total number of images analysed was 1886.

\subsection{Preprocessing}
The image preprocessing involved conversion from archival TIFF to JPEG and resizing them to a maximum horizontal resolution of 1600 pixels, maintaining the original aspect ratio. This was done to reduce the AI inference’s computational load. The resizing was performed using the Lanczos filter of the Pillow library.

\subsection{AI inference}
The AI models used were the following:
\begin{itemize}
    \item OpenAI GPT: 4.1-nano, 4.1-mini, 4.1, 5.
    \item Gemini: 2.5 Flash-Lite.
\end{itemize}

The OpenAI GPT models \cite{radford2018improving, radford2019language, brown2020language, achiam2023gpt} and the Gemini models \cite{team2023gemini,team2024gemini,comanici2025gemini} are large language models based on the Transformer architecture \cite{vaswani2017attention}. This architecture uses the concept of attention \cite{galassi2020attention} to contextualise the use of words in sentences.
While large language models are models that are just designed to predict next words in sentences, since the transformer development they have quickly dominated all natural language processing tasks \cite{zhao2023survey,yang2024harnessing}. Originally developed to be used with language, in recent years their use has been extended to images \cite{dosovitskiy2020image,han2022survey}, allowing the development of multimodal LLMs able to use images and text as inputs. This multimodality allowed them to be used for tasks such as the one implemented in this project. The Gemini models are based on the mixture-of-experts transformer architecture \cite{shazeer2017outrageously,fedus2022switch,riquelme2021scaling}.

For the complete set of images, the number of equivalent tokens processed was: 7.2M (GPT-4.1-nano), 4.7M (GPT-4.1-mini), 2M (GPT-4.1), 1.7M (GPT-5), and 2.9M (Gemini 2.5 Flash-Lite).

The prompts used for the models consisted of 3 strings concatenated together. The first part defined the AI persona, the data, and the task to be carried out. There were 2 versions of this first part (called ``multiple'' and ``single''), designed to use either the full set of record images or just the first image containing the title page. The second part of the prompt included hints given to the models to facilitate the task. There were 2 versions of this second part (called ``simple'' and ``advanced''), including or not references to the use of authoritative names and catalogue records. The third part of the prompt was used to specify the output format (``MARC'', ``JSON'', ``BIBFRAME''). In total, considering all possible combinations, there were 12 variations of the prompts.

A total of 60 experiments were carried out, applying for each of the 5 models the 12 versions of the prompts. After applying the 60 experiments to each of the 87 records, we obtained a total of 5220 outputs.

\subsection{Quantitative Evaluation}

\subsubsection{Ground Truth}

To evaluate the results of the experiments, we used the manually produced catalogue cards for each dissertation. Each of the dissertation folders also contained a scanned image of the corresponding card. The extracted content of the card was compared with the outputs produced by the AI models from the dissertation page images.

Extracting card content used a similar AI-driven process to the one described above for the dissertations as whole. However, this is a much simpler task since a card only contains human-curated catalogue information, and it is already largely segregated by the card layout. This use-case primarily tests handwriting recognition and the sorting of data into the correct fields, for a sample with very neat handwriting and easily distinguishable data characteristics (title, author, data, language and place). AI tools were expected to (and did) perform well. The OpenAI GPT-4.1 model was applied to the card images to extract their output and generate the different formats for use in the evaluation. 

\subsubsection{Evaluation }

To evaluate the correspondence between the ground truth and the generated outputs, we used 3 metrics:

\begin{itemize}
    \item Jaccard similarity.
    \item Semantic similarity.
    \item BLEU score.
\end{itemize}

The Jaccard similarity \cite{niwattanakul2013using} evaluates the overlap between the sets of words of the ground truth and the output. It is calculated as the number of words in the intersection of the sets divided by the number of words in the union of the sets.

The semantic similarity \cite{chandrasekaran2021evolution} was implemented using the SentenceTransformer library\footnote{\url{https://sbert.net/}} \cite{reimers-2019-sentence-bert} with the  `all-MiniLM-L6-v2' model to calculate the embedding representation of ground truth and output, and then calculate the cosine product between both as the final metric score. The embedding representation of a text is a numerical representation of a text produced using an AI model. Each numerical representation can be understood as a vector. The cosine similarity evaluates how similar the vectors are by evaluating how different the directions they point to are. The more different the texts, the more their numerical vector representations will diverge. Unlike the previous metrics, which evaluate the similarity between literal words, these vectors produce a semantic representation of the texts as a whole, independently of the words used.

The BLEU score \cite{papineni2002bleu} was implemented using the sacreBLEU library\footnote{\url{https://github.com/mjpost/sacrebleu}} \cite{post2018call}. This score computes a similarity score based on N-gram precision and output length.

Using these methods, we are able to evaluate more than 5000 outputs produced, which would have been impossible to do using a completely manual evaluation.

In addition to these main evaluations, we conducted a brief analysis using a subset of 13 dissertations that had been recently re-catalogued more thoroughly. While this only evaluates part of the full 87 record dataset and one of the output formats (MARC), this comparison is against what could be considered a ``gold-standard" human-sourced record.

\subsection{Qualitative Evaluation }

The data fields from the experiments were also extracted from their respective formats and loaded into Excel spreadsheets for visual inspection. While the quantitative analysis can identify broad performance characteristics of the algorithms under test, it does not provide an insight into the \textit{nature} of the variance between results, and will statistically suppress occasional ``hallucinations". This part of the evaluation therefore focuses on the character and consistency of variations rather than outright performance at scale as an indication of the nature of any intervention or monitoring that might be necessary.

\section{Results}

The experiments carried out have been evaluated both quantitatively and qualitatively. We present these results below.

\subsection{Quantitative Evaluation}

The results of the experiments are presented in Tables \ref{tab:marc}, \ref{tab:json}, and \ref{tab:bibframe}.

\begin{table}[!ht]
    \centering
    \setlength{\tabcolsep}{3pt}
    \caption{Results using \textbf{MARC} format}
    \begin{tabular}{llllll}
    \hline
        Model & prompt & images & Jaccard & Sem. & BLEU \\ \hline
        GPT-4.1 & advanced & multiple & 0.192 & 0.793 & 0.485 \\ 
        GPT-4.1 & advanced & single & 0.182 & 0.652 & 0.476 \\ 
        GPT-4.1-mini & advanced & multiple & 0.324 & 0.794 & 0.738 \\ 
        GPT-4.1-mini & advanced & single & 0.336 & 0.807 & 0.749 \\ 
        GPT-4.1-nano & advanced & multiple & 0.249 & 0.787 & 0.650 \\ 
        GPT-4.1-nano & advanced & single & 0.256 & 0.771 & 0.671 \\ 
        GPT-5 & advanced & multiple & 0.322 & 0.804 & 0.681 \\ 
        GPT-5 & advanced & single & 0.327 & 0.802 & 0.712 \\ 
        GPT-4.1 & simple & multiple & 0.250 & 0.816 & 0.577 \\ 
        GPT-4.1 & simple & single & 0.247 & 0.802 & 0.582 \\ 
        GPT-4.1-mini & simple & multiple & 0.326 & 0.811 & 0.742 \\ 
        GPT-4.1-mini & simple & single & 0.341 & 0.823 & 0.755 \\ 
        GPT-4.1-nano & simple & multiple & 0.259 & 0.799 & 0.647 \\ 
        GPT-4.1-nano & simple & single & 0.285 & 0.816 & 0.690 \\ 
        GPT-5 & simple & multiple & 0.311 & 0.813 & 0.693 \\ 
        GPT-5 & simple & single & 0.301 & 0.823 & 0.640 \\ 
        Gemini-2.5-FL & advanced & multiple & 0.265 & 0.796 & 0.611 \\ 
        Gemini-2.5-FL & advanced & single & 0.230 & 0.767 & 0.537 \\ 
        Gemini-2.5-FL & simple & multiple & 0.274 & 0.812 & 0.619 \\ 
        Gemini-2.5-FL & simple & single & 0.245 & 0.798 & 0.600 \\ \hline
    \end{tabular}
    \label{tab:marc}
\end{table}

\begin{table}[!ht]
    \centering
    \setlength{\tabcolsep}{3pt}
    \caption{Results using \textbf{JSON} format}
    \begin{tabular}{llllll}
    \hline
        Model & prompt & images & Jaccard & Sem. & BLEU \\ \hline
        GPT-4.1 & advanced & multiple & 0.138 & 0.722 & 0.156 \\ 
        GPT-4.1 & advanced & single & 0.149 & 0.705 & 0.169 \\ 
        GPT-4.1-mini & advanced & multiple & 0.350 & 0.837 & 0.527 \\ 
        GPT-4.1-mini & advanced & single & 0.349 & 0.744 & 0.532 \\ 
        GPT-4.1-nano & advanced & multiple & 0.285 & 0.796 & 0.464 \\ 
        GPT-4.1-nano & advanced & single & 0.298 & 0.785 & 0.484 \\ 
        GPT-5 & advanced & multiple & 0.345 & 0.823 & 0.493 \\ 
        GPT-5 & advanced & single & 0.353 & 0.847 & 0.523 \\ 
        GPT-4.1 & simple & multiple & 0.279 & 0.762 & 0.383 \\ 
        GPT-4.1 & simple & single & 0.312 & 0.760 & 0.445 \\ 
        GPT-4.1-mini & simple & multiple & 0.337 & 0.821 & 0.485 \\ 
        GPT-4.1-mini & simple & single & 0.356 & 0.721 & 0.528 \\ 
        GPT-4.1-nano & simple & multiple & 0.285 & 0.800 & 0.449 \\ 
        GPT-4.1-nano & simple & single & 0.303 & 0.800 & 0.483 \\ 
        GPT-5 & simple & multiple & 0.329 & 0.816 & 0.468 \\ 
        GPT-5 & simple & single & 0.334 & 0.825 & 0.487 \\ 
        Gemini-2.5-FL & advanced & multiple & 0.315 & 0.717 & 0.469 \\ 
        Gemini-2.5-FL & advanced & single & 0.314 & 0.682 & 0.479 \\ 
        Gemini-2.5-FL & simple & multiple & 0.306 & 0.718 & 0.437 \\ 
        Gemini-2.5-FL & simple & single & 0.303 & 0.690 & 0.482 \\ \hline
    \end{tabular}
    \label{tab:json}
\end{table}

\begin{table}[!ht]
    \centering
    \setlength{\tabcolsep}{3pt}
    \caption{Results using \textbf{BIBFRAME} format}
    \begin{tabular}{llllll}
    \hline
        Model & prompt & images & Jaccard & Sem. & BLEU \\ \hline
        GPT-4.1 & advanced & multiple & 0.183 & 0.601 & 0.456 \\ 
        GPT-4.1 & advanced & single & 0.173 & 0.407 & 0.478 \\ 
        GPT-4.1-mini & advanced & multiple & 0.206 & 0.837 & 0.591 \\ 
        GPT-4.1-mini & advanced & single & 0.235 & 0.860 & 0.614 \\ 
        GPT-4.1-nano & advanced & multiple & 0.147 & 0.818 & 0.438 \\ 
        GPT-4.1-nano & advanced & single & 0.173 & 0.823 & 0.462 \\ 
        GPT-5 & advanced & multiple & 0.249 & 0.807 & 0.541 \\ 
        GPT-5 & advanced & single & 0.252 & 0.816 & 0.567 \\ 
        GPT-4.1 & simple & multiple & 0.202 & 0.662 & 0.486 \\ 
        GPT-4.1 & simple & single & 0.199 & 0.465 & 0.518 \\ 
        GPT-4.1-mini & simple & multiple & 0.209 & 0.882 & 0.602 \\ 
        GPT-4.1-mini & simple & single & 0.244 & 0.896 & 0.615 \\ 
        GPT-4.1-nano & simple & multiple & 0.140 & 0.809 & 0.419 \\ 
        GPT-4.1-nano & simple & single & 0.172 & 0.845 & 0.453 \\ 
        GPT-5 & simple & multiple & 0.264 & 0.850 & 0.540 \\ 
        GPT-5 & simple & single & 0.262 & 0.846 & 0.545 \\ 
        Gemini-2.5-FL & advanced & multiple & 0.046 & 0.600 & 0.188 \\ 
        Gemini-2.5-FL & advanced & single & 0.087 & 0.686 & 0.308 \\ 
        Gemini-2.5-FL & simple & multiple & 0.025 & 0.597 & 0.125 \\ 
        Gemini-2.5-FL & simple & single & 0.075 & 0.621 & 0.242 \\ \hline
    \end{tabular}
    \label{tab:bibframe}
\end{table}

Additionally, we present the average of the previous results averaged by model, in Tables \ref{tab:marc_model_avg}, \ref{tab:json_model_avg}, and \ref{tab:bibframe_model_avg}.

\begin{table}[!ht]
    \centering
    \caption{Results using \textbf{MARC} format - Average by model}
    \begin{tabular}{llll}
    \hline
        ~ & Jaccard & Semantic & BLEU \\  \hline
        GPT-4.1 & 0.218 & 0.766 & 0.530 \\ 
        GPT-4.1-mini & \textbf{0.332} & 0.809 & \textbf{0.746} \\ 
        GPT-4.1-nano & 0.262 & 0.793 & 0.664 \\ 
        GPT-5 & 0.315 & \textbf{0.810} & 0.681 \\ 
        Gemini-2.5-FL & 0.253 & 0.793 & 0.592 \\ \hline
    \end{tabular}
    \label{tab:marc_model_avg}
\end{table}

\begin{table}[!ht]
    \centering
    \caption{Results using \textbf{JSON} format - Average by model}
    \begin{tabular}{llll}
    \hline
        ~ & Jaccard & Semantic & BLEU \\  \hline
        GPT-4.1 & 0.219 & 0.737 & 0.288 \\ 
        GPT-4.1-mini & \textbf{0.348} & 0.781 & \textbf{0.518} \\ 
        GPT-4.1-nano & 0.293 & 0.795 & 0.470 \\ 
        GPT-5 & 0.340 & \textbf{0.828} & 0.493 \\ 
        Gemini-2.5-FL & 0.309 & 0.702 & 0.467 \\ \hline
    \end{tabular}
    \label{tab:json_model_avg}
\end{table}

\begin{table}[!ht]
    \centering
    \caption{Results using \textbf{BIBFRAME} format - Average by model}
    \begin{tabular}{llll}
    \hline
        ~ & Jaccard & Semantic & BLEU \\  \hline
        GPT-4.1 & 0.189 & 0.534 & 0.484 \\ 
        GPT-4.1-mini & 0.223 & \textbf{0.869} & \textbf{0.605} \\ 
        GPT-4.1-nano & 0.158 & 0.824 & 0.443 \\ 
        GPT-5 & \textbf{0.257} & 0.830 & 0.548 \\ 
        Gemini-2.5-FL & 0.058 & 0.626 & 0.216 \\ \hline
    \end{tabular}
    \label{tab:bibframe_model_avg}
\end{table}

It can be seen that GPT-4.1-mini obtains the top results in most of the categories, followed by GPT-5. In particular, the latter seems to obtain better results for the semantic metric, while the former produces better results for Jaccard and BLEU.

In Tables \ref{tab:marc_prompt_avg}, \ref{tab:json_prompt_avg}, and \ref{tab:bibframe_prompt_avg} are presented the results averaged by prompt category.

\begin{table}[!ht]
    \centering
    \caption{Results using \textbf{MARC} format - Average by prompt}
    \begin{tabular}{llll}
    \hline
        ~ & Jaccard & Semantic & BLEU \\  \hline
        advanced & 0.268 & 0.777 & 0.631 \\ 
        simple & \textbf{0.284}& \textbf{0.811} & \textbf{0.654} \\  \hline
        multiple & \textbf{0.277} & \textbf{0.802} & \textbf{0.644} \\ 
        single & 0.275 & 0.786 & 0.641 \\ \hline
    \end{tabular}
    \label{tab:marc_prompt_avg}
\end{table}

\begin{table}[!ht]
    \centering
    \caption{Results using \textbf{JSON} format - Average by prompt}
    \begin{tabular}{llll}
    \hline
        ~ & Jaccard & Semantic & BLEU \\  \hline
        advanced & 0.290 & 0.766 & 0.430 \\ 
        simple & \textbf{0.314} & \textbf{0.771} & \textbf{0.465} \\  \hline
        multiple & 0.297 &\textbf{0.781} & 0.433 \\ 
        single & \textbf{0.307} & 0.756 & \textbf{0.461} \\ \hline
    \end{tabular}
    \label{tab:json_prompt_avg}
\end{table}

\begin{table}[!ht]
    \centering
    \caption{Results using \textbf{BIBFRAME} format - Average by prompt}
    \begin{tabular}{llll}
    \hline
        ~ & Jaccard & Semantic & BLEU \\  \hline
        advanced & 0.175 & 0.725 & \textbf{0.464} \\ 
        simple & \textbf{0.179} & \textbf{0.747} & 0.454 \\  \hline
        multiple & 0.167 & \textbf{0.746} & 0.439 \\ 
        single & \textbf{0.187} & 0.726 & \textbf{0.480} \\ \hline
    \end{tabular}
    \label{tab:bibframe_prompt_avg}
\end{table}

The ``simple'' version of the prompt is, in most of the cases, producing better results than the ``advanced'' one. The single-image case works better than using all the images, for BIBFRAME and JSON, and in any case presents very close results to the multiple-image case, while drastically reducing the computational cost.

Looking at the output formats, the three formats produce very similar results with respect to the semantic metric. Regarding Jaccard similarity, MARC and JSON produce better results than BIBFRAME. Considering the BLEU score, MARC obtains better results than JSON and BIBFRAME. In summary, MARC seems to produce the best results, followed by JSON, and then by BIBFRAME.

As we mentioned in the previous section, we conducted an additional evaluation using a subset of 13 manually created MARC records. Those evaluation results are presented in Table \ref{tab:marc_second_experiment}.

\begin{table}[!ht]
    \centering
    \setlength{\tabcolsep}{3pt}
    \caption{Results using \textbf{MARC} format and the human-created records}
    \begin{tabular}{llllll}
    \hline
        Model & prompt & images & Jaccard & Sem. & BLEU \\ \hline
        GPT-4.1 & advanced & multiple & 0.144 & 0.562 & 0.157 \\ 
        GPT-4.1 & advanced & single & 0.132 & 0.577 & 0.132 \\ 
        GPT-4.1-mini & advanced & multiple & 0.111 & 0.552 & 0.092 \\ 
        GPT-4.1-mini & advanced & single & 0.134 & 0.578 & 0.115 \\ 
        GPT-4.1-nano & advanced & multiple & 0.096 & 0.518 & 0.072 \\ 
        GPT-4.1-nano & advanced & single & 0.120 & 0.515 & 0.094 \\ 
        GPT-5 & advanced & multiple & 0.129 & 0.552 & 0.122 \\ 
        GPT-5 & advanced & single & 0.138 & 0.573 & 0.112 \\ 
        GPT-4.1 & simple & multiple & 0.127 & 0.589 & 0.131 \\ 
        GPT-4.1 & simple & single & 0.139 & 0.558 & 0.133 \\ 
        GPT-4.1-mini & simple & multiple & 0.111 & 0.561 & 0.100 \\ 
        GPT-4.1-mini & simple & single & 0.138 & 0.567 & 0.113 \\ 
        GPT-4.1-nano & simple & multiple & 0.090 & 0.448 & 0.072 \\ 
        GPT-4.1-nano & simple & single & 0.093 & 0.502 & 0.080 \\ 
        GPT-5 & simple & multiple & 0.112 & 0.564 & 0.097 \\ 
        GPT-5 & simple & single & 0.107 & 0.508 & 0.082 \\ 
        Gemini-2.5-FL & advanced & multiple & 0.108 & 0.548 & 0.093 \\ 
        Gemini-2.5-FL & advanced & single & 0.110 & 0.489 & 0.118 \\ 
        Gemini-2.5-FL & simple & multiple & 0.105 & 0.583 & 0.100 \\ 
        Gemini-2.5-FL & simple & single & 0.118 & 0.598 & 0.116 \\ \hline
    \end{tabular}
    \label{tab:marc_second_experiment}
\end{table}

In this evaluation, GPT-4.1 obtains the best results, followed closely by GPT-4.1-mini for the Jaccard and semantic metrics, and with a larger difference for the BLEU metric.

The ``advanced'' version of the prompt obtains a better result for Jaccard and BLEU, but not for the semantic metric. The same happens for the single-image cases.

\subsection{Qualitative Evaluation}

Manual inspection of the outputs of the various experiments identified a number of characteristic idiosyncrasies for the various AI tools that appeared regardless of the model used. These are noteworthy in that they are not generally the type of behaviour that would be expected of a human cataloguer. We did not encounter ``hallucinations", where the model deviated completely from the task at hand, but more subtle variations which are harder to detect as part of a quality assurance process.

Any single record from the test dataset exhibits many of these behaviours. As an illustration, we will present extracts from the outputs of the following experiments, which involve the same item (shelfmark: Diss. D 1,2), same input image (title page only), same prompt (simple instructions), same model (GPT-4.1), but different output formats requested (MARCXML, JSON and BIBFRAME). This has \textbf{not} been specially chosen to illustrate these points but was simply the first item processed in our tests. 

\subsubsection{MARCXML} Abbreviated extract from full record 

\scriptsize
\begin{lstlisting}
  <datafield tag="100" ind1="1" ind2=" ">
    <subfield code="a">
      Fagel, Francois,
    </subfield>
    <subfield code="e">
      author.
    </subfield>
  </datafield>
  <datafield tag="245" ind1="1" ind2="0">
    <subfield code="a">
      Specimen juris publici inaugurale de guarantia 
      foederum,
    </subfield>
  </datafield>
  <datafield tag="300" ind1=" " ind2=" ">
    <subfield code="a">
      1 volume (unpaged)
    </subfield>
    <subfield code="c">
      22 cm
    </subfield>
  </datafield>
  <datafield tag="502" ind1=" " ind2=" ">
    <subfield code="b">
      Doctor of Laws (Juris Utriusque Doctor),
    </subfield>
    <subfield code="c">
      Leiden University,
    </subfield>
    <subfield code="d">
      1759
    </subfield>
  </datafield>
  <datafield tag="710" ind1="2" ind2=" ">
    <subfield code="a">
      Universiteit Leiden.
    </subfield>
    <subfield code="e">
      degree granting institution.
    </subfield>
  </datafield>
\end{lstlisting}\normalsize
\subsubsection{JSON} Full record 
\scriptsize
\begin{lstlisting}
{
  "Title": "Specimen Juris Publici Inaugurale 
    de Guarantia Foederum",
  "Author": "Francois Fagel",
  "Date of Publication": "1759",
  "Place of Publication": "Lugduni Batavorum",
  "Language": "Latin",
  "Degree": 
    "Doctoratus in Utriusque Jure (Doctor of Law)",
  "Degree granting institution": 
    "Academia Lugduno-Batava (Leiden University)",
  "Subject Classification": 
    "International law -- Treaties; 
    Law -- Netherlands -- History -- 18th century"
}
\end{lstlisting}
\normalsize

\subsubsection{BIBFRAME} Abbreviated extract from full record 
\scriptsize
\begin{lstlisting}
<bf:Person>
  <rdfs:label>Francois Fagel</rdfs:label>
  <bf:label>Fagel, Francois</bf:label>
</bf:Person>
  
<!-- Work entity for the dissertation -->
<bf:Work>
  <bf:title><bf:Title>
    <bf:mainTitle>Specimen juris publici inaugurale
      de Guarantia Foederum</bf:mainTitle>
  </bf:Title></bf:title>
  <bf:contribution><bf:Contribution>
    <bf:agent rdf:resource=
      "http://example.org/person/fagel-francois"/>
    <bf:role><bf:Role>
      <rdfs:label>author</rdfs:label>
    </bf:Role></bf:role>
  </bf:Contribution></bf:contribution>
  <bf:subject><bf:Topic rdf:about=
    "http://id.loc.gov/authorities/subjects/sh85130611">
    <rdfs:label>Treaties</rdfs:label>
  </bf:Topic></bf:subject>
  <bf:subject><bf:Topic rdf:about=
    "http://id.loc.gov/authorities/subjects/sh85191227">
    <rdfs:label>International law</rdfs:label>
  </bf:Topic></bf:subject>

  <!-- Degree information -->
  <bf:dissertation><bf:Dissertation>
    <bf:degree><bf:Degree>
      <rdfs:label>Doctor of Law 
        (Juris Doctor, Doctoratus)</rdfs:label>
    </bf:Degree></bf:degree>
      <bf:grantingInstitution>
        <bf:Organization rdf:about=
          "http://id.loc.gov/authorities/names/n79018173">
          <rdfs:label>Universitas Lugduno-Batava 
            (Leiden University)</rdfs:label>
        </bf:Organization>
      </bf:grantingInstitution>
  </bf:Dissertation></bf:dissertation>
</bf:Work>

<!-- Instance entity for the physical dissertation -->
<bf:Instance>
  <bf:instanceOf 
    rdf:resource=
      "http://example.org/work/fagel-guarantia-foederum"/>
  <bf:provisionActivity><bf:ProvisionActivity>
    <bf:place><bf:Place>
      <rdfs:label>Lugduni Batavorum 
        [Leiden, Netherlands]</rdfs:label>
    </bf:Place></bf:place>
    <bf:date>1770</bf:date>
  </bf:ProvisionActivity></bf:provisionActivity>
</bf:Instance>
\end{lstlisting}
\normalsize

\subsubsection{Non-intuitive Prompt Sensitivity} By comparing these records we can see that changing the output format has a non-intuitive effect on the way that information is extracted from the page and formatted. In MARCXML, the dissertation title is rendered in lower case with an initial capital, in JSON the major words are capitalised and in BIBFRAME only the primary topic is capitalised. In the original, the entire title is printed in upper case. MARCXML data also seems to include trailing punctuation for some reason.   

\subsubsection{Data Synthesis} In the MARCXML record, datafield tag 300 corresponds to physical information about the original object. The GPT model has synthesised a plausible but completely fictitious dimension of 22cm since the digitised images submitted for scanning have had the scale ruler cropped out. The prompt did not include physical information in the required MARCXML fields.

In the BIBFRAME record, all the authoritative links to Library of Congress resources were either fictitious (https://id.loc.gov/authorities/subjects/sh85191227.html) or incorrect (http://id.loc.gov/authorities/names/n79018173 refers to a person rather than Leiden University), although of the correct form to pass casual inspection.

\subsubsection{Date ``Blindness"} In the BIBFRAME record, the change in required output format causes the algorithm to fail to recognise the date correctly, generating 1770 rather than the (correct) value of 1759 seen in the MARCXML and JSON outputs. Incorrect extraction of dates seemed to be particularly prevalent.   
\subsubsection{Translation} In addition to failures, the models also exhibited some emergent behaviours that were not part of the prompted tasks, but might be potentially useful or counterproductive depending on circumstances. The source documents were primarily in Latin but, in many cases, the algorithms helpfully expanded and translated terms into modern languages, albeit somewhat at random. In the source document for the records above, the academic institution involved is referenced by the terms ``Lugduni Batvorum" and ``Acad. Lugd. Bat." which was correctly interpreted as Leiden University in all three cases. However, the MARCXML record also references ``Universiteit Lieden" in one place - which is linguistically appropriate for the institution but confusing given that English is used elsewhere. In BIBFRAME we see the incorrect resynthesised pseudo-Latinate ``Universitas Lugduno-Batava", whereas ``Academia Lugduno-Batava (Leiden University)" in the JSON records, expands the Latin abbreviation correctly. 

\subsubsection{Inconsistency}An overriding observation about these behaviours is that they do not appear consistently, as can be seen from analysing just a single record in the test dataset. A prompt/output format that performs well for one item may not be optimal for another, ostensibly similar, item. The plausibility of many of these variant results makes error detection difficult.

Another type of inconsistency encountered in the project is the tendency for the algorithms to stray from the tasks specified in the prompt. Data is falsified in an attempt to generate fuller MARCXML and BIBFRAME records despite prompts that indicate precisely which fields are required. Conversely, unrequested translations revealed a potentially useful capability.      

\section{Discussion}

The qualitative and quantitative assessments of the experiments described above have led to the following points for discussion and recommendations regarding the use of AI models.

\subsection{Larger Models and More Data is not Always Better}
The project tested several models from OpenAI as well as some from other vendors. The quantitative comparison of the outputs of the various algorithms found that GPT-4.1-mini performed marginally better than either GPT-4.1 or GPT-5.1. Secondly, there is no benefit in using all the scanned images instead of just the title page, and in some cases doing so makes the results worse.  

\textit{Larger and newer models do not necessarily perform better depending on use-case. Smaller models are considerably cheaper, and more environmentally-friendly, than larger models. There is no guarantee that newer models will perform better or as well as older models. This should be considered when building workflows that are dependent on particular models or tools.}

\subsection{Complex Tasks}
 
Generating modern digital records is more complex than simple transcription, involving the use of authoritative references, controlled vocabularies, etc., which proved too complex for a single prompt despite being well within notional context windows for the algorithms concerned.  

\textit{Complex tasks need to be broken down into simpler tasks, some of which may be suitable for automation using automated AI (or other) tools. This is probably best accomplished by an expert in the field, in conjunction with a business analyst.}      

\subsection{Emergent Behaviours and Hallucinations}
 
Tools could randomly perform additional tasks that were not specified in the prompt. In some cases, these were “hallucinations” such as synthesising physical data where none existed to conform to the algorithm's interpretation of record completeness. At other times, it revealed a potentially useful capability, such as translating the Latin name of a degree-awarding institution into modern English or Dutch.  

\textit{Identifying such cases and redefining tasks and prompts to minimise “hallucinations” and leverage helpful emergent behaviours is an essential part of operationalising tool use.    } 

\subsection{Prompt Sensitivity}
 
“Prompt engineering” is more of an art than a science, based on empirical evidence of what seems to work rather than a deep understanding of the internal mechanics of LLMs. Crafting effective prompts thus requires a certain amount of trial-and-error with realistic test datasets to get optimal performance. Our testing revealed that performance is not only dependent on prompt construction in readily understandable ways – for example, overly complex task specifications are not always helpful (more complex and detailed prompting showed improvements in some metrics and regressions in others) – but also in unexpected ways – changing the required output format for metadata causes the model to handle dates less reliably. In quantitative terms, MARC seems to produce the best results, followed by JSON, and then by BIBFRAME. At an individual record level, changing the output format has a significant non-intuitive impact on how data is extracted from the page images and processed into metadata records.  

\textit{A prompt that works well for one model does not necessarily work well for another model, so any time a model is changed or upgraded, a re-engineering effort is required to ensure performance is not compromised. This is a potentially non-trivial cost because of rapid model evolution.}

\subsection{Inconsistency }
Both manual and AI processes are likely to result in a certain level of error. However, the nature of the errors generated by LLM-based tools does differ qualitatively from human errors in that they are less consistent and logical. Whereas human errors are frequently systematic, the appearance of hallucinations and helpful emergent behaviours can vary on a record-by-record basis, which makes error checking and correction more difficult either by manual or algorithmic means.

\textit{Since correction is relatively expensive, the goal in operationalising LLM-based tools should be to first minimise the inconsistency in output to acceptable levels.}

\section{Conclusions}
This workstream was undertaken as a research activity to build in-house understanding of LLM-based tools. 
The use case was processing digitised documents and catalogue cards with a view to automatically generating catalogue records suitable for use in Library Systems. This revealed a number of characteristics of these AI tools which should be borne in mind when considering their operational implementation.  

In addition to the qualitative analysis of the results and the recommendations developed based on it, in this article we have presented a quantitative analysis of the experiments carried out, which underpin these recommendations and highlight practical aspects in the implementation of LLMs in use cases similar to that presented in this project.

Overall, LLM-based tools showed promise but require careful, use-case specific implementations assisted by domain experts, the acceptance of an unavoidable level of inconsistency/inaccuracy and the expectation that ongoing maintenance will be required. 

\bibliographystyle{SageV}

\bibliography{biblio.bib}

\end{document}